\documentclass[lettersize,journal]{IEEEtran}
\usepackage{amsmath,amsfonts}
\usepackage{algorithm}
\usepackage{array}
\usepackage[caption=false,font=normalsize,labelfont=sf,textfont=sf]{subfig}
\usepackage{textcomp}
\usepackage{stfloats}
\usepackage{url}
\usepackage{verbatim}
\usepackage{graphicx}
\usepackage{cite}
\usepackage{gensymb}
\usepackage{multirow}
\usepackage{xcolor}
\usepackage{algpseudocode}
\usepackage{algorithm}
\begin{document}

\title{Automotive RADAR sub-sampling via object detection networks: Leveraging prior signal information}


\author{Madhumitha Sakthi, Ahmed Tewfik, Marius Arvinte, Haris Vikalo
\thanks{The authors are with the Department of Electrical and Computer Engineering, The University of Texas at Austin. E-mails: {madhumithasakthi.iyer@utexas.edu, tewfik@austin.utexas.edu, arvinte@utexas.edu, hvikalo@ece.utexas.edu.}}
}



\maketitle

\begin{abstract} Automotive radar has increasingly attracted attention due to growing interest in autonomous driving technologies. Acquiring situational awareness using multimodal data collected at high sampling rates by various sensing devices including cameras, LiDAR, and radar requires considerable power, memory and compute resources which are often limited at an edge device. 
In this paper, we present a novel adaptive radar sub-sampling algorithm designed to identify regions that require more detailed/accurate reconstruction based on prior environmental conditions' knowledge, enabling near-optimal performance at considerably lower effective sampling rates.
Designed to robustly perform under variable weather conditions, the algorithm was shown on the Oxford raw radar and RADIATE dataset to achieve accurate reconstruction utilizing only 10\% of the original samples in good weather and 20\% in extreme (snow, fog) weather conditions.
A further modification of the algorithm incorporates object motion to enable reliable identification of important regions. 
This includes monitoring possible future occlusions caused by objects detected in the present frame.
Finally, we train a YOLO network on the RADIATE dataset to perform object detection directly on RADAR data and obtain a 6.6\% AP50 improvement over the baseline Faster R-CNN network. 
\end{abstract}

\begin{IEEEkeywords}
Compressive sensing, automotive radar, sub-sampling, Faster R-CNN, YOLO, object detection, measurement matrix, signal acquisition
\end{IEEEkeywords}

\section{Introduction}
\IEEEPARstart{A}{utomotive} RADAR is a robust all-weather sensor providing assistance to autonomous driving systems. 
While other sensing technologies commonly used by autonomous vehicles, including cameras and LiDAR, are generally capable of providing situational awareness, they struggle under extreme weather conditions (e.g., heavy rain, fog and snow) \cite{rad-cam-fusion}. Radar, however, is a robust all-weather sensor that can provide both depth and locations of the objects in the surroundings. For instance, broadly used Frequency-Modulated Continuous Wave (FMCW) radar operates in the wide $30$-$300$ GHz band to provide highly accurate range data \cite{ROD-Net}. Prior research focused on utilizing radar to assist images \cite{centre-fusion,radar1,radar2,rrpn}, where the raw radar data is processed to extract point clouds which are subsequently used for object detection. More recently, methods relying solely on raw radar data have been shown to achieve robust object detection, making it a particularly attractive choice for acquiring situational awareness in extreme weather conditions \cite{ROD-Net, track-detect-radar, efficient-ROD, ROD-squeeze-excite, radiate-object-detection, ROD-dimension-apart-nw, ana-radar-object-detection}. 

To achieve desired target accuracy, radars typically collect data at very high rates. For instance, the aforementioned FMCW radar \cite{radiate} operating at 76-77 GHz generates
data at about 16 Gbps. 
The analog-to-digital converter (ADC) thus needs to function at an extremely high rate to capture the reflected signal and forward it to the DSP for processing \cite{ADC-limitations}. The time-domain signal is then converted to the frequency domain, where the frequency is proportional to the range of the objects found in a frame. To relax the computational and energy requirements on the data processing pipeline, and meet latency constraints, it would be beneficial to reduce the data rate without compromising quality of the acquired situational awareness. To this end, previous studies have explored the use of compressed sensing (CS) methods to enable trade-off between sampling rate and acquisition quality\cite{cs-candes}.

In this paper, we present a method for adaptive sub-sampling of raw radar data via compressed sensing aided by the results of object detection from previous frames. In particular, the proposed algorithm splits each radar frame into uniform-sized blocks and relies on object detection to determine the blocks that require more attention. Given a total sampling budget, the algorithm solves an aptly posed linear program to adaptively allocate a higher sampling rate to important blocks and a lower sampling rate to the remaining ones. 
Essentially, the algorithm would aid in increasing the effective output resolution of the ADC in regions (i.e., range bins) where an object is expected to be present (as reflected by the expected time delay of the measurement) while decreasing the effective ADC output resolution in regions where an object is not expected to be found. 
The azimuth is also obtained utilizing prior information about the location of the objects of interest.  

The proposed framework assumes the general availability of both camera and radar data and adapts the sampling strategy to varied weather conditions. In good weather, the sub-sampling algorithm uses the previous frame's data (images, radar) to decide importance of regions in the incoming radar data; the object detection algorithm runs on the images and identifies the bounding boxes and object class. If an object is either missed by the object detection network or is in the blind spot of a camera, previous radar data and images are used to assist in identifying regions of interest. After converting the detection result from image coordinates to radar coordinates, a linear program (LP) is formed and solved to determine the exact sampling rate for each radar block. 
In bad/extreme weather conditions, the algorithm does not rely on image sensors to determine regions of interest but rather uses as prior information the positions of tracked objects predicted using a Kalman filter. 

The main contributions of the paper are as follows:
\begin{itemize}
    \item A novel adaptive radar sub-sampling algorithm designed to select regions that require more attention, leading to near-optimal reconstruction performance at a considerably lower sampling rate.
    \item A methodology for addressing possible occlusions in future frames by the objects detected in the current frame. 
    \item A sub-sampling technique that relies on object motion prediction to improve the selection of important regions.
    \item A design that promotes robustness across varied weather conditions. 
    \item An experimental verification via a YOLO object detection network trained on RADIATE \cite{radiate}, demonstrating a significant improvement (in terms of AP50 metric) over the baseline Faster R-CNN network (63.8\% vs. 57.2\% AP50). 
\end{itemize}

\begin{figure*}[ht!]
\begin{center}
\includegraphics[width=0.8\linewidth]{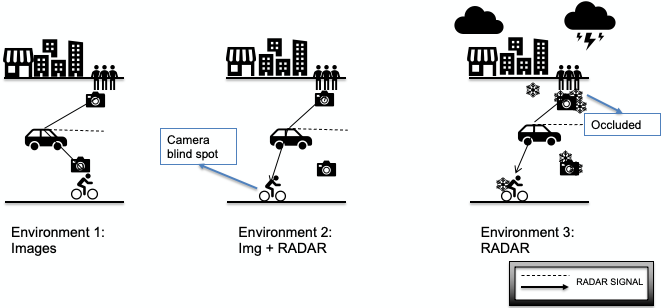}
\centering
\caption{\label{Overall-ALGO-Illus}Illustration of various sub-sampling scenarios. In Environment 1, only the information from images is used while in Environment 2, both image and radar information is used for further sub-sampling. In Environment 3 with extreme weather conditions, only the previous radar information is used.}
\end{center}
\vspace{-6mm}
\end{figure*}

Figure \ref{Overall-ALGO-Illus} shows the environments where the proposed radar sub-sampling algorithms could be suitable. In Environment 1, where both camera and radar data are available, we use the previous image information to guide the current radar frame acquisition. In Environment 2, where the camera coverage is either missing or unreliable, previous images as well as previous radar data is utilized to guide the acquisition. Finally, in Environment 3, where objects could be occluded due to bad weather conditions, only the previous radar signals is used for current radar frame acquisition. 

The remainder of the paper is organized as follows. Section II provides an overview of the existing efficient acquisition methods followed by the object detection networks trained on images and radar point cloud and radar-based object detection. The proposed algorithms are presented in Section III, along with details on the compressed sensing procedure, the LP optimization problem, and the object detection network utilized by the algorithm. Section IV presents experimental results on the Oxford robocar dataset \cite{oxford} and RADIATE \cite{radiate} datasets. The paper is concluded in Section V. 

\section{Related work}
\subsection{Compressed sensing for radar systems}
The compressed sensing (CS) techniques enable the acquisition of data at sub-Nyquist sampling rates without a loss of information or striking a trade-off between sampling rate and reconstruction quality. Applying CS ideas, \cite{CS-SAR-Radar} presents reliable reconstruction of Synthetic Aperture Radar (SAR) data acquired at a 70\% sampling rate. In \cite{radar-cs2}, CS is utilized for noisy radar data reconstruction sampled at a 30\% rate, while \cite{radar-cs-40} presents efficient CS-based reconstruction of a frequency-modulated continuous wave (FMCW) radar sampled at 40\% of the Nyquist rate. 
In \cite{radar-cs3}, the authors analyzed various CS reconstruction algorithms including Orthogonal Matching Pursuit (OMP) and Basis Pursuit De-noising (BPDN), and have shown that, in automotive settings, OMP has superior performance. Our proposed framework relies on the Basis Pursuit (BP) algorithm for signal reconstruction since BP, though more computationally expensive, requires fewer measurements than OMP \cite{OmpVsBp}. 

Adaptive CS is a broadly used technique for increasing the sampling rate in regions deemed important \cite{adaptive-radar-cs1,adaptive-radar-cs2,adaptive-radar-cs3}. The scheme in \cite{adaptive-radar-cs1} utilizes previously received pulse interval and relies on the constant false alarm rate (CFAR) to decide how to undersample the pulsed radar data. In \cite{adaptive-radar-cs2}, an adaptive CS algorithm was used to improve target tracking performance in static settings.  Furthermore, the authors of \cite{adaptive-radar-cs3} aimed at optimizing the measurement matrix where only the targets are moving, improving performance at the cost of increasing computational complexity. In contrast, we apply an adaptive CS algorithm for radar acquisition in settings where the objects and the vehicle deploying the sensors are potentially moving.
Moreover, a distinctive feature of our work is that the measurement matrix size increases for certain regions of the environment as we allocate them a larger sampling budget while maintaining the overall sampling budget and reconstruction complexity. In another related work \cite{ROI-lidar}, the acquisition of LiDAR data is guided using the Region-of-Interest information determined based on the results of image segmentation; our work helps guide radar data acquisition utilizing the results of 2-D object detection. 

The ADC's sampling frequency and resolution play a major role in the acquisition of the intermediate frequency (IF) signal following the signal mixing stage. In order to limit the rate of the received radar data, the authors of \cite{BiMIMO} used a bit-limited MIMO radar with an additional analog filter to form a global hybrid analog-digital system. In \cite{ont-bit-radar}, high-resolution ADCs were replaced by one-bit ADCs using time-varying thresholds. In order to estimate the angle and Doppler frequency from one-bit sampled data, the \cite{Angle-doppler} uses a maximum-likelihood-based method. In contrast, our method relies on DNN output to assist in efficient spatial sampling, thereby focusing on the regions of interest while limiting the overall bit budget during acquisition without the need for analog filters or threshold-modified ADC to acquire the range-azimuth radar data.

When compressively sampling signals, the measurement matrix plays an important role in ensuring that the signal can be reconstructed. In addition to often used Gaussian matrices, binary measurement matrices received considerable attention due to their hardware advantages. In \cite{simp-meas-matrix}, the authors proposed a Binary Permuted Block Diagonal (BPBD) measurement matrix with equally-sized diagonal blocks permuted along the columns to create randomness. There, BPBD was compared with the scrambled Fourier and Partial Noiselet alternatives, among others, showing comparable reconstruction performance on images. In related work, \cite{bin-image-BCS-SPL} proposed a CS reconstruction approach that relies on an extended smoothed-projected Landweber algorithm. In \cite{CS-ECG-binary-meas-matrix}, binary random measurement matrices were used for CS of ECG signal, while in \cite{detr-bin-matrix-ecg} binary block diagonal matrix without permutation was introduced as a deterministic measurement matrix to compress and recover electrocardiogram (ECG) and Electromyography signals. In \cite{binary-matrix-radar-ground}, authors applied a binary measurement matrix to the problem of investigating soils or stone walls, showing successful reconstruction with reduced sampling rates, while \cite{parv08} used a binary measurement matrix to facilitate compressive measurements in DNA microarrays. Evidently, the binary measurement matrix has been successful in a variety of CS applications. In this paper, we implement the BPD matrix for automotive radar acquisition and demonstrate their object detection performance in terms of AP and AP50 metrics. To the best of our knowledge, this is the first work to experiment with BPD matrices for radar acquisition in automotive settings and show their effectiveness using task-relevant metrics such as AP and AP50. 

\subsection{Object detection using radar and images}
Previous studies demonstrated the advantage of combining radar data with images for improved object detection performance. In \cite{radar4}, a spatial attention fusion method was proposed; the method was embedded at the feature-extracting stage to effectively leverage both radar and vision features. In \cite{rad_img1}, the authors deployed Faster R-CNN object detection network, replacing the selective search-based region proposal algorithm with the region proposals generated using radar data points. This work experimented on the NuScenes \cite{nuscenes} dataset, using selective search to show an AP improvement from 41.8\% to 43.0\%. The authors in \cite{radar1} report more accurate 3-D object detection performance using the combination of radar and images as compared to LiDAR and images. In \cite{centre-fusion}, the authors used a center point detection network and proposed a frustum-based method to facilitate 3-D object detection, showing improved occluded object detection.
In \cite{radar4}, radar and image features combined via spatial attention to improve the object detection performance while using the FCOS \cite{FCOS} object detection network pipeline. 

\subsection{Object detection using radar}
This line of research focuses on generating object detection networks from radar data without relying on any other sensors. To this end, several raw radar datasets with object annotations have been collected and open-sourced \cite{radiate, ROD-Net}. In \cite{ROD-Net}, the authors 
proposed a ROD-Net network for radar object detection that resulted in 86\% average precision; they also released a dataset named CRUW containing object annotations on RF images. The method in \cite{track-detect-radar} combined particle filter-based tracking and object detection to enable radar-based object identification, demonstrating accurate performance while alternating between tracking and detection to reduce the computational load incurred by real-time processing. In \cite{radiate-object-detection}, a channel boosting feature ensemble method was reported and validated on the radiate dataset; the method uses a transformer neural network for direct object detection from radar frames. However, this approach requires bulky backbone networks such as resnet-50 or resnet-101 to process RGB, LUV and LAB radar frames individually, rendering its practical applications challenging.

\begin{figure*}[ht!]
\begin{center}
\includegraphics[width=\linewidth]{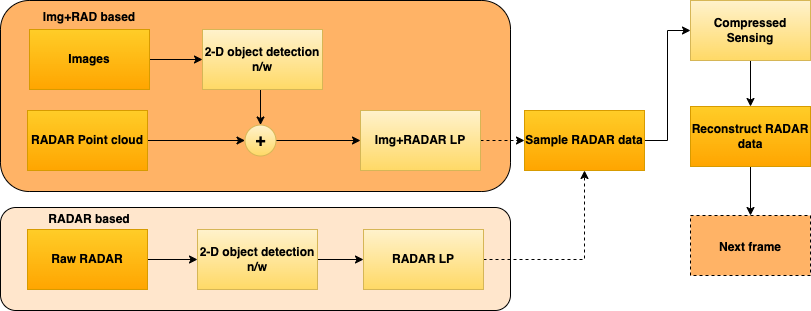}
\centering
\caption{\label{Algorithm}The overall radar sub-sampling algorithm. The first block shows the pipeline that uses both images and radar point clouds for important region determination. The second block is utilized in extreme weather conditions by only using the previous raw radar data.}
\label{Algo}
\end{center}
\vspace{-6mm}
\end{figure*}

\section{Method}
The three environments illustrated in Figure \ref{Overall-ALGO-Illus} present distinct challenges and thus require different algorithms for radar signal acquisition. Figure \ref{Algorithm} shows a detailed block diagram of various inputs required and the corresponding processing pipeline. In all cases, an object detection network is used and the radar data is acquired via compressed sensing. However, the input signal for each algorithm and the associate linear programming (LP) problem is defined independently. The details of each pipeline are explained in the following sections. 

\subsection{CompRADIMG - Sub-sampling using previous images and radar data}
We use adaptive compressed sensing to split the measured frame into multiple blocks and probe each block with a predetermined number of measurements. In particular, $m$ measurements are collected from each block. This procedure is predicated on two assumptions: the signal being measured is sparse in some domain and the measurement matrix follows restricted isometry property \cite{Binary-CS-Image}. We specifically assume that the data is sparse in the Discrete Cosine Transform (DCT) domain. 
Given the original signal $x \in R^n$, each block is acquired with a random BPD measurement matrix $\phi \in R^{mxn}$ resulting in measurements $y \in R^m$. During decompression, the signal is recovered by running the basis pursuit algorithm to solve 
\begin{equation}
\begin{matrix}
    \underset{x}{\min} \|\theta x\|_1 \\
    {\mbox s.t. } \;\; \phi x = y,
\end{matrix}
\end{equation}
where $\theta$ denotes the DCT transformation matrix \cite{image-adaptive-cs}. The baseline in our experiments is the sampling framework where the samples/bits are uniformly distributed across the frame being sampled without using any prior frames information. Another baseline is the CFAR algorithm wherein the important regions are determined based on the CFAR results on the previous raw radar data; this information is utilized to dynamically allocate the sampling budget across the frame. 

In good weather conditions, as shown in figure \ref{Algorithm}, our proposed algorithm CompRADIMG performs RADAR sub-sampling using both prior image and RADAR data. The results are shown on the Oxford raw radar dataset \cite{oxford}. 
The data, captured at 4Hz frame rate, has range resolution 4.38cm and azimuth resolution 0.9$^{\circ}$, with the total range of 163m with 360$^{\circ}$ horizontal field-of-view (HFoV). The rear camera has 180$^{\circ}$ HFoV while the front camera has 66$^{\circ}$ at a given timestamp, with a blind spot of 57$^{\circ}$ on either side of the vehicle. In this setup, the images from the camera were captured at $t=0$s and processed by the Faster R-CNN object detection network. The output from the network that arrives at $t=0.12$s is converted from camera coordinates to radar azimuth blocks. The LP algorithm is then used to dynamically determine the sampling budget for various radar blocks; this sampling rate is used to acquire radar data at $t=0.25$s. The procedure repeats for the upcoming frames. 


If our radar sub-sampling algorithm relies only on prior image data for sub-sampling \cite{cs-object}, in cases where an object detection network  misses an object or if the object appears in the blind spot of the camera, the objects could be missed while sub-sampling the radar data. This could be detrimental to further system processing that would rely on the RADAR data for depth estimation or path planning and so on. In order to address this scenario and improve the overall sub-sampling algorithm's performance, we introduce our CompRADIMG algorithm which combines the previous image and radar data to identify the regions with objects of interest. The bounding boxes of objects from the images in the image coordinate are converted to azimuth in the radar frame. After determining important azimuth regions, the CFAR algorithm on the previous radar data directly identifies blocks containing objects. This is then provided as an additional important block to the LP algorithm which is used to determine the sampling rates for all blocks of the present radar frame.


In order to adaptively increase the sampling rate for certain regions while maintaining the overall sampling budget, 
the radar data was split into bins of size 25x100, each having 22.5$^{\circ}$ azimuth and 4.38m range accounting for the average object in a scene (pedestrians, cars, truck). In total, there were 37 range blocks and 16 azimuth blocks. The LP algorithm below was used to determine the dynamic sampling rate for each radar block. 
Here, $a_1$ denotes blocks with pedestrians or bicycles, $a_2$ denotes blocks with a car, and $a_3$ denotes all other regions. $a_1, a_2$ and $a_3$ correspond to three azimuth blocks of decreasing importance. The first 18 range blocks (78.84m) correspond to $r_1$ radar region while the following 19 range blocks (83.22m) constitute $r_2$ region. The sampling rates for the blocks are optimized and are denoted by $x_1, x_2, x_3, x_4$. Given the total radar data of $S$, the 10\% sampling rate is $0.1S$. Furthermore, 
$b_1$ denotes the number of blocks identified as important based on the previous radar frame using the CFAR algorithm. $b_2$ and $b_3$ indicate the number of blocks to be removed from the other regions. Therefore, as $b_1$ increases, $b_2$ and $b_3$ are used to remove blocks from other important regions if they were already marked important based on the coordinates from the image detection result. This ensures that a region is not allocated a sampling budget twice.  The CFAR algorithm is run on each row (range) with 300 training cells, 50 guard cells and 1e-3 false alarm rate. For a given frame, the sampling rate across azimuth blocks is optimized using the total sampling budget of less than 10\% with the below objective function. Since regions with small objects would benefit the most from having a higher sampling budget, $x1$ is constrained as thrice as big as the least important region, while the next best sampling rate $x2$ is twice as big. Finally, the region starting from 78.84m away from the autonomous vehicle is allocated the least sampling rate of less than 2.5\%, identified as $x4$. 
The LP problem is as follows:

For any vector $x\in \mathbb{R}^4$, let 
\begin{align*}
    f(x) &= (a_1r_1+b_1)x_1 + (a_2r_1+b_2)x_2 \\
    &~~~~~+ (a_3r_1+b_3)x_3 + (a_1 + a_2 + a_3)r_2x_4
\end{align*}
Then, we have the following linear program
\begin{align*}
\max_{x\geq 0} &\; f(x)\\
s.t. &\; x_1 - 3x_3 = 0,\; x_2 - 2x_3 = 0\\
& b_1+b_2+b_3 = 0\\
& f(x) \leq 0.1S, \;0.02 \leq x_4 \leq 0.025,\\
& 0.05\leq x_i\leq 0.4 ~~\text{for } i = 1,2,3\\
\end{align*}

The pseudocode for our CompRADIMG is presented in Algorithm \ref{alg:cap}. The function $f_{NN}$ is the pretrained object detection network that predicts the bounding box on input camera images. The function $f_{CFAR}$ \cite{radarFunda} determines important blocks across every row in the radar frame $\hat{X_t}$ which is included as an important block $b_1$ in the LP problem and the corresponding blocks are removed using $b_2$ and $b_3$ in the other regions. We determine the approximate azimuth of an object from its bounding box. Since we divide the acquisition region into 16 azimuth blocks, we use a simple formula to relate the pixel location based on the field of view of the camera. $ f_A(B) = (centre_x/ X_{max} - X_{min}) *(\theta_{max} - \theta_{min})$ where $centre_x$ is the  x-coordinate value of the detected object bounding box's centre, $\theta_{min}, \theta_{max}$ are the minimum and maximum field of view of the camera relative to the position of the camera in the birds eye view frame and $X_{min}, X_{max}$ is typically 0 and the total width of the image. The sampling rate is determined using the above-mentioned LP problem and is presented as $f_{LP}$ in Algorithm \ref{alg:cap}.
Finally, the frame is compressed and reconstructed using the compressed sensing algorithm $f_{CS}$. In the case of a 10\% sampling rate, this results in compressed radar data with 10\% of the total samples present in the input radar frame. 

\begin{algorithm}
\caption{CompRADIMG Algorithm}\label{alg:cap}
Input: RADAR frame $X \in \mathbb{R}^{m \times n}$  and image frame $I$ \\
Output: Reconstructed RADAR frame $\hat{X}$ \\
    At $ t = 1$ , Initialize first radar frame $\hat{X_t}$

    \textbf{for} $t = 2, \dots, T$ \textbf{do} 
    \begin{enumerate}
    \item Determine object bounding boxes $B \leftarrow f_{NN}(I_{t-1})$ 
    \item Determine important image blocks $b \leftarrow f_{CFAR}(\hat{X}_{t-1})$ 
    \item Determine important azimuth $\hat{A} \leftarrow f_{A}(B)$ 
    \item Determine sampling rate for $t$-th frame $X_{t}$ using $x_1,x_2,x_3,x_4 \leftarrow f_{LP}(\hat{A},b)$ 
    \item Output compressed and reconstructed $\hat{X_{t}} \leftarrow f_{CS}(x_1,x_2,x_3,x_4, X_t)$
\end{enumerate}
\textbf{end for}

\end{algorithm}
\begin{figure*}[ht!]
\begin{center}
\includegraphics[width=\linewidth]{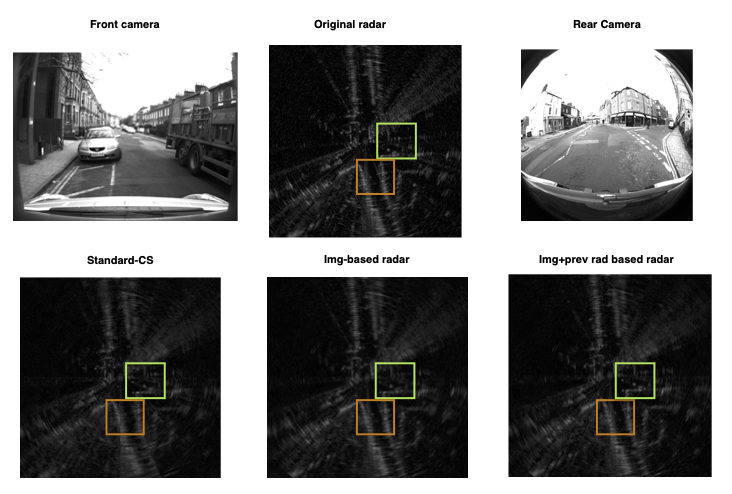}
\centering
\caption{In scene 2, frame 9, the front camera, rear camera, original radar and reconstructions are shown. The green box indicates the car to the rear right. The car was missed by the image-based radar algorithm but is reconstructed by the CompRADIMG algorithm. The orange box indicates the bicycle to the rear. The bicycle was captured by the proposed algorithm but missed by the baseline. \textit{Best viewed as a digital copy by zooming in.} }
\label{scene2-frame9}
\end{center}
\vspace{-6mm}
\end{figure*}
\subsection{Analysis of measurement matrices}
In compressed sensing, the measurement matrix is used to take measurements of the incoming signal and is used to compress the signal based on the sampling rate. Typically, a Gaussian measurement matrix is used and the linear combination of the measured signal is taken and transmitted. However, this Gaussian measurement matrix requires float multiplication of the incoming signal. The authors in \cite{simp-meas-matrix} successfully replaced the Gaussian measurement matrix with Binary Permuted Block Diagonal (BPBD) measurement matrix and showed similar performance to the Gaussian matrices. The BPBD measurement matrices had only binary elements and the need for float multiplications was avoided. Given the sparsity of the measured RADAR data, we extended the BPBD as a Binary Permuted Diagonal (BPD) measurement matrix\cite{cs-object}. Therefore, instead of blocks of measurements from the radar signal that gets added, in our BPD matrix, we directly measure the exact radar signal, avoiding the addition as well. Therefore, power-consuming complex multipliers used in the case of Gaussian matrix are avoided and can now be replaced with hardware-efficient elements such as switches and selectors in the case of Binary matrix \cite{bin-image-BCS-SPL}. Additionally, we show the performance of the measurement matrices in terms of AP and AP50, task-relevant metrics as opposed to the traditional PSNR.


\subsection{RADAR-based object detection}
The authors in \cite{radiate} proposed a Faster R-CNN object detection network for predicting the bounding boxes of vehicles directly on the radar frame. The network was trained with ResNet50 backbone and Feature Pyramid Networks (FPN) to predict class and bounding box coordinates. The model predicted either a vehicle class or no-object. The radiate data with 8 different classes were combined to form a single vehicle class and ignored pedestrian annotations similar to the model proposed by the authors \cite{radiate}. For this task of direct object detection on the raw radar data, we propose to use the YOLOv5 object detection network.  In Faster R-CNN, the features from the images are first processed through a region proposal network and then the generated proposals are used for the final bounding box and class prediction. Whereas, the YOLO network does not have a region proposal step and directly predicts from the image split into $S \times S$ cells. Therefore, it has the advantage of higher inference speed. However, the model we trained has the advantage of lesser parameters and hence lesser computational resource utilization.  

\subsection{CompRPD: RADAR sub-sampling using previous RADAR}
In adverse weather conditions, the sub-sampling algorithm is designed to only rely on previous radar data for the identification of important regions in the present frame.
Extending the work in \cite{EUSIPCO2022}, we present a detect-and-predict algorithm for important region identification. In the previous work, the radar data was split into equal-sized blocks and based on the detection result, the objects of interest's locations were identified. However, in this approach, given the algorithm is designed to run for multiple frames (20 frames) together, we used another vital information, the predicted location of the objects based on their previous locations. 

We used the Kalman filter to predict the location of the object in the present frame based on the detection result from the previous frame. The first anchor frame, sampled at a higher sampling rate is processed through the object detection network. The bounding boxes of the object's location in the first frame are used to initialize the Kalman filter. In the second frame, the predicted object's bounding boxes from the current frame are matched to the previous frame's bounding boxes prediction from the Kalman filter using IOUs and each mapped object is then used to predict and update the Kalman filter for that particular object. This would continue until the next initialization point or the Kalman filter has reached the maximum age hyper-parameter due to a lack of bounding box matches. Once the confidence of the Kalman filter's predictions is determined based on the age of the tracked frames, the predicted bounding boxes are used to determine the location of the objects in the upcoming frame that is being sampled. In all the other cases, the detected object's bounding boxes are directly used for sampling. Therefore, in this improved algorithm, the knowledge of the object's future position is used for sub-sampling in addition to the current position. Therefore, once the objects are detected in the previous frame, either the same coordinates are used or the predicted coordinates by the Kalman filter are used. These coordinates are then converted into radar polar coordinates and the blocks with the objects of interest are identified. 
In this proposed algorithm, an inverted T-block sampling pattern is used, where only adjacent polar blocks that are away from the autonomous vehicle are chosen while considering additional polar blocks which are close to the vehicle.
This was done to ensure that sufficient blocks beside the autonomous vehicles were sampled for safely operating at a lower sampling budget. Moreover, during the Cartesian to polar block conversion, the blocks close to the vehicle translate to a lower area and hence more blocks are needed for sampling a larger area. 
Following this, the LP algorithm \cite{cs-object} is used to dynamically determine the sampling rate for each block in the present radar frame. 
We tested this algorithm on the RADIATE dataset \cite{radiate} which was captured every 0.25s, 4Hz frame rate. Compared to the Oxford dataset, this dataset had lower resolution and lower range. Therefore, we split the polar domain radar frame into 18\degree and 8.4m in range as $20 \times 48$ sized blocks from the entire $400 \times 576$ frame. 

For any vector $x\in \mathbb{R}^2$, let 
\begin{align*}
    f(x) &= I.w.h.x_1 + O.w.h.x_2.
\end{align*}
Then, we have the following linear program
\begin{align*}
\max_{x\geq 0} &\; f(x)\\
s.t. &\; x_1 >= 1.1x_2\\
& f(x) \leq S, \;x_1l \leq x_1 \leq x_1u,\\
& x_2l\leq x_2\leq x_2u. \\
\end{align*}
The width of the block, 48 (range) is denoted by $w$ and the height of 20 (azimuth) is denoted by $h$. The total number of important blocks is represented by $I$ and the total number of other blocks is denoted by $O$. There are a total of 240 blocks derived by splitting the radar image of size $400 \times 576$ into equal-sized blocks of size $20 \times 48$. The sampling rate for important and other blocks is given by $x_1$ and $x_2$ respectively. The condition, $f(x) < S$ is used to limit the number of samples -- e.g., to 10\% or 20\% or 30\% of the total samples (400x576). Also, the condition $x_1 >= 1.1x_2$ aids in ensuring that the sampling rate for the important regions is higher than for the other regions. In the case of the three sampling rate cases, 10\%, 20\%, and 30\%, the lower bound for $x_1$ is 0.1, 0.2, and 0.3 while the upper bound is 0.55. 
The upper bound was determined such that the reconstruction matches that of the original and the lower bounds were chosen to ensure that the sampling rate is at least as in the standard-CS case. In the case of $x_2$, the upper bounds were set to 0.1, 0.2 and 0.3 for 10\%, 20\% and 30\% sampling rates, respectively, since the number of measurements could be limited due to the lack of the object of interest in the other regions and the lower bound was set to 0.07 to ensure there are enough samples to support reconstruction. Therefore, once the sampling rates were determined by solving the LP algorithm, they can be used for the subsequent radar frame and the reconstructed radar is used anew for object detection; this, in turn, is further used for important region identification and the loop continues for 20 frames. Again, the 21\textsuperscript{st} frame is considered as an anchor frame and the entire area is sampled at a 40\% sampling rate in order to avoid the bias of focusing on only certain regions and missing an object present in the non-important region. 
In the case of 20\% sampling rate, S is set to 46080 (20\% of $400 \times 576$) and these measurements are adaptively allocated across the important and other regions of the radar frame based on the LP results while maintaining the overall sampling budget to be within 20\% or 46080 measurements for 20\% sampled radar data.

\begin{algorithm}
\caption{CompRPD Algorithm}\label{alg:RPD}
Input: RADAR frame $X \in \mathbb{R}^{m \times n}$ \\
Output: Reconstructed RADAR frame $\hat{X}$ \\
For $t = 1$, initialize first radar frame $\hat{X_t}$ \\
\textbf{for} $t = 2, \dots, T$ \textbf{do}
\begin{enumerate}
    \item Determine object bounding boxes $B \leftarrow f_{NN}(X_{t-1})$
    \item \textbf{if} $t-1$ in $fullRate$ \textbf{then}
    \item \hskip1.5em Initialize filter $K_{1,\dots,len(B)}$ with $NNPred_{BB}$
    \item \hskip1.5em $Final_{BB} \leftarrow B$
    \item \textbf{else}
    \item \hskip1.5em Determine Kalman predictions $KPred_{BB} \leftarrow Pred(K_{t-1})$
    \item \hskip1.5em $IOU = f_{IOU}(KPred_{BB},NNPred_{BB})$
    \item \hskip1.5em \textbf{if} $IOU > 0$ \textbf{then}
    \item \hskip2.5em Refine $K$ using $NNPred_{BB}$
    \item \hskip2.5em Update filter $K$ using $NNPred_{BB}$
    \item \hskip2.5em Update $K_{age}$ by 1
    \item \hskip2.5em $Final_{BB} = K_{age} > MinAge \ ? \ KPred_{BB} : NNPred_{BB}$
    \item \hskip1.5em \textbf{end if}
    \item \textbf{end if}
    \item Determine polar coordinates $G \leftarrow f_{CToP}(Final_{BB})$
    \item Determine sampling grids $I \leftarrow f_{Grids}(G)$
    \item Set all other grids to $O$
    \item Determine sampling rate for $t$-th frame $X_{t}$ using $x_1,x_2 \leftarrow f_{LP}(I,O)$
    \item Output compressed and reconstructed $\hat{X_{t}} \leftarrow f_{CS}(x_1,x_2, X_t)$
    
\end{enumerate}
\textbf{end for} 
\end{algorithm}

The pseudo-code for the CompRPD algorithm is given in Algorithm \ref{alg:RPD}. The function $f_{NN}$ is the pretrained object detection network on radar data which directly predicts object bounding boxes on input radar Cartesian images. The $fullRate$ is used to define the frames at which the radar frame would be sampled uniformly at a fixed maximum sampling rate. Therefore, at the frame right after full frame, the bounding boxes from the previous frame would be used to initialize the Kalman filters and the sampling is performed using the current predictions directly. In the case of subsequent frames, the Kalman predictions are used instead of the NN predictions, if the age of the filter is greater than a fixed hyper-parameter. The function $f_{CToP}$ is used to convert the bounding boxes in Cartesian to polar coordinates and the central grid for each prediction is determined. Depending on the size and the location of the object, the surround grids are marked as important using the function $f_{Grids}$ while all the other grids are set as $O$. The above-mentioned LP, $f_{LP}$ is used to determine the sampling rate for every frame for the important and other grids and the computed sampling rate is used for compression and reconstruction of the current radar frame using the function $F_{CS}$.

\subsection{Multi-rate sampling using anchor frames}
Given the acquisition system runs on the fly, it is important to determine the frequency of the anchor frame: the frame acquired at the maximum allowed sampling rate for safety. Therefore, we also analyzed the rate and the ideal frequency at which the anchor frames should be sampled. We showed in \cite{EUSIPCO2022}, an algorithm where the first frame is acquired at full rate while the following 20 frames are acquired at the user-defined sampling rate of 10\%, 20\% or 30\%. As a trade-off, the data acquired at a 30\% sampling rate exhibited the closest object detection performance to original raw radar data. In the case of 20 frames, propagation error was not an issue either. However, in this paper, we explore and present to the users with cases of having anchor frames every 10 frames and 5 frames instead of 20 frames 
to best use the limited sampling budget while extracting the maximum possible information about the environment.  
In real-time at every anchor frame, once the IF signal is derived from the reflected signal, the ADC could be used to sample the signal at a fixed lower resolution, in this case, 40\% sampling rate i.e., about 3.2 bits/sample across the entire frame in order to gain unbiased information about the frame irrespective of the previous frames' information.

\section{Experiment}
\subsection{Data}
In the experiments, we use the Oxford radar robocar dataset \cite{oxford} and the RADIATE dataset \cite{radiate}. The Oxford dataset consists of Navtech CTS350-X Millimeter-Wave FMCW radar data along with the camera data from Point Grey Bumblebee XB3 rear camera, Point Grey Grasshopper2 stereo front camera and other sensor data. The data is collected from a total of 280km roads in Oxford, UK, with the front camera capturing at 16Hz frames per second (FPS), the rear camera capturing at 17Hz, and the radar capturing at 4Hz. To the best of our knowledge, the Oxford dataset is the only publicly available raw radar dataset with both front and rear camera data. Since object annotations were not provided for the radar data, we needed to manually annotate a subset of it, and ended up testing our algorithm on three random scenes with about 11 frames each. 

The RADIATE dataset \cite{radiate} was collected in extreme weather conditions including snow, fog, rain and so on. The dataset contains radar, LiDAR, camera and GPS data. The radar data was collected using Navtech CTS350-X with a 360\degree Horizontal Field of View (HFoV) and 100-meter range at 4Hz. This resulted in range-azimuth images of size 400x576 where the rows represent the angle and the column represents the range. The authors of \cite{radiate} released 300 hours of annotated radar data, for both good and bad weather conditions, with annotations on the Cartesian radar images. Similar to \cite{radiate}, in our experiments we classify objects as vehicles or background by defining the vehicle class as either car, bus, bicycle, truck, van or motorbike. To test our proposed algorithm, we selected from the RADIATE dataset 40 frames for each snow, fog, motorway, city and night conditions. 


\subsection{CompRADIMG: Subsampling using image and radar}
We tested the radar subsampling using our image and radar-based algorithm on the oxford \cite{oxford} dataset. We randomly took three scenes each with about 11 frames. In the first scene, the vehicle was driven on a straight road with pedestrians on either side of the walkway. A van was following the vehicle and at a distance to the front, a bus was coming towards the vehicle and a pedestrian was crossing at the front. In frame 2, the person to the right was missed by the first algorithm but was captured by the second algorithm because the image and radar-based algorithm, deemed the region to the right of the blind spot important based on radar CFAR detection. However, in frame 6, the person was missed, possibly due to a lesser available sampling budget after allocating the budget to additional areas marked by the CFAR algorithm. 

In the second scene, the car was passing through an intersection with a parked car to the front left and a truck passing by to the right. At the rear of the vehicle, there were multiple bicycles (bikes) at a distance and a pedestrian crossing to the rear left. In a few frames, the cars parked to the right vehicle appear at the camera blind spot. In Figure \ref{scene2-frame9}, we show frame 9's radar and camera images. Especially in this scene, the car parked to the right side of the vehicle was not reconstructed by the image-based radar since it appeared on the cameras' blind spot. Whereas, in the image and radar-based algorithm, this car was reconstructed. In frames 6,7,8,9 of scene 2, this car was reconstructed in all frames except frame 7 by our proposed algorithm. In Figure \ref{scene2-frame9}, the front camera, original radar, and rear camera along with reconstructed data are presented. The orange box highlights the bicycle that is visible on the original radar but was not reconstructed by the standard CS algorithm. However, the proposed algorithms reconstructed the bicycle. The green box highlights the car that was present to the right of the autonomous vehicle. This was reconstructed by the CompRADIMG algorithm but was missed by the image-based and the standard-CS algorithm.  

In the third scene, the vehicle was passing a crowded intersection with a bus to the front and buses to the rear left. There were multiple pedestrians on either side of the road with a few bicycles as well. The proposed algorithm reconstructed the person to the right in frames 2,6,7 while it was missed by the image-based algorithm. But, in frame 4, the person to the right was missed by our CompRADIMG algorithm. Finally, the bicycle to the right was reconstructed by the proposed algorithm while it was missed by the image-based algorithm and standard-CS methods. 

\begin{table}[H]
\vspace{-4mm}
\caption{\label{results}
S-CS denotes standard CS, A-1\cite{cs-object} is the algorithm with only images and A-2 is the algorithm with images and radar (CompRADIMG). The presence of an object is indicated as 'yes' and if the object is faint or absent, it is indicated as 'no'. F-Front, Re-Rear, FL-Front Left.}
\centering
\begin{tabular}{|c| c| c| c| c| c| c|} 
 \hline
 Scene & Frame &Object & S-CS & CFAR & A-1  & A-2 \\ [0.5ex] 
 \hline\hline
 \multirow{3}{4em}{Scene1} &  2 & Person (R)  & no & yes & no & yes \\
 &  3 & Person(FL)  & no & no & yes&  yes \\
 &  6 & Person (R)  & no & yes & yes&  no \\
 &  9 & Person (R)  & no & no & yes&  yes \\
 \hline
 \multirow{3}{4em}{Scene2} &  2-5,8,9,11 & Bike (Re)  & no & no & yes & yes \\
 &  6,8 & Car (R) & no & no& no & yes \\
 &  9 & Car (R) & no & yes & no & yes \\
 &  7 & Car (R) & no & no & yes&  no \\
 \hline
 \multirow{2}{4em}{Scene3} &  2,6 & Person (R) & no & yes & no & yes \\
 &  4 & Car (R) & no & no & yes & yes \\
 &  4 & Person (R) & no & no &yes & no \\
 &  7 & Person (R) & no & no & no & yes \\
 &  10 & Bike (R) & no & no & no & yes \\
 \hline
\end{tabular}
\vspace{-4mm}
\end{table}

\subsection{Measurement matrix analysis}
Previously \cite{cs-object}, the BPD matrix for automotive radar data acquisition was proposed and the reconstruction quality was measured in PSNR. Typically, we found that BPD has a similar PSNR compared to Gaussian and BPBD matrix. However, since radar object annotations are available in the case of the RADIATE dataset, we could test the measurement matrix performance in terms of AP and AP50 metrics that are more relevant to the task compared to PSNR. We reconstructed 5 scenes of radiate data with 40 frames each using standard-CS algorithm at 10\%, 20\% and 30\% sampling rates using the three measurement matrices and evaluated using both Faster R-CNN and YOLOv5l network. The results are shown in table \ref{meas-matrix-result-OD}. In the case of the AP and AP50 metrics, BPD, the proposed measurement matrix was better than Gaussian and BPBD. Especially the difference in performance is more prominent at lower sampling rates than at higher sampling rates. YOLOv5 network's performance was better than Faster R-CNN in the case of 30\% sampling rate while in the other cases, Faster R-CNN was better. In general binary matrices, BPBD and BPD are better than Gaussian matrices as well. Therefore, it is vital to evaluate using task-relevant metrics such as AP or AP50 in addition to PSNR to conclude the best measurement matrix for automotive radar acquisition using the CS algorithm.   
\begin{table}[H]
\caption{\label{meas-matrix-result-OD} The scenes were reconstructed using the standard CS algorithm and tested using Faster R-CNN and YOLOv5 network. The AP and AP50 metrics are reported.}
\centering
\begin{tabular}{|c| c| c| c| c|} 
 \hline
 Sampling rate & Network & Gaussian &  BPBD & BPD \\ [0.5ex] 
 \hline\hline
 \multirow{3}{4em}{10\%} & Faster R-CNN &5.5/2.2 &5.6/2.0 & 10.0/3.6\\
 & Yolov5 &3.1/1.4 &3.7/1.6 & 7.1 /2.4\\
 \hline
 \multirow{3}{4em}{20\%} &   Faster R-CNN &31.0/12.3 & 32.3/12.6& 37.7/15.0\\
 & Yolov5 &27.6/11.2 &29.3/11.8 & 36.3/14.2\\
 \hline
 \multirow{2}{4em}{30\%} &   Faster R-CNN &47.8/19.3 & 48.2/19.8& 49.2/20.3\\
 & Yolov5 &50.5/20.8 &50.4/20.9 & 54.3/22.0\\
 \hline
\end{tabular}
\vspace{-2mm}
\end{table}

\begin{table}[H]
\caption{\label{meas-matrix-result} The scenes were reconstructed using the standard CS algorithm and tested using Faster R-CNN and YOLOv5 network. The PSNR metric is reported.}
\centering
\begin{tabular}{|c| c| c| c|} 
 \hline
 Sampling rate  & Gaussian &  BPBD & BPD \\ [0.5ex] 
 \hline\hline
10\%  & 25.39 & 25.65 & 25.40\\
 \hline
 20\%  & 26.58 & 26.68 & 26.59\\
 \hline
 30\% & 27.66 & 27.71 & 27.66\\
 \hline
\end{tabular}
\end{table}

\subsection{RADAR-based object detection}
The Faster R-CNN network was trained by using the recommended parameters by \cite{radiate}. We used the ResNet-50 backbone and feature pyramidal network and trained for 90,000 iterations with 128 images per batch and 0.00025 learning rate. The network was modified for a single class classification of vehicles, a combination of all released annotations except for pedestrians and group\_of\_pedestrians. The Yolo networks \cite{ultralytics} were again trained for a single class - vehicle prediction using radar frames of size 1280x1280. The Yolo5l network was trained for 10 epochs with batch\_size 2. Finally, Yolo5l has 46M parameters, and inference on an NVIDIA 2070 Super GPU took 70ms/img. Similarly, the Yolo5m network was also trained for 10 epochs with batch\_size 2 and the network had 20M parameters and inference speed was 41ms/img. Across the three models that were trained on the RADIATE dataset, Yolo5l gave the best AP50 and AP but at a higher number of parameters and inference time. Also, Yolo5m took a little more than half the inference time of Yolo5l, it has half the number of parameters for 61.9 AP50 on the overall dataset. Also, to the best of our knowledge, we are the first ones to train a YOLO network for the RADIATE task and show significant performance improvement over Faster R-CNN networks that were previously used in the literature. 
\begin{table}[H]
\caption{\label{sampling-OD}
Radar object detection results [AP/AP50].}
\centering
\begin{tabular}{|c| c| c| c| c|} 
 \hline
 Model & AP50 & mAP & Params & Inference [ms] \\ [0.5ex] 
 \hline\hline
 Faster R-CNN[1] & 57.2 & 22.6 & 41M  &  63ms/img\\  
 \hline
 Yolov5m & 61.9  & 26.4  & 21M & 41ms/img \\ 
 \hline
 Yolov5l & 63.8  & 27.0  & 46M & 70ms/img \\ 
 \hline
\end{tabular}
\end{table}
\begin{figure*}[ht!]
\begin{center}
\includegraphics[width=\linewidth]{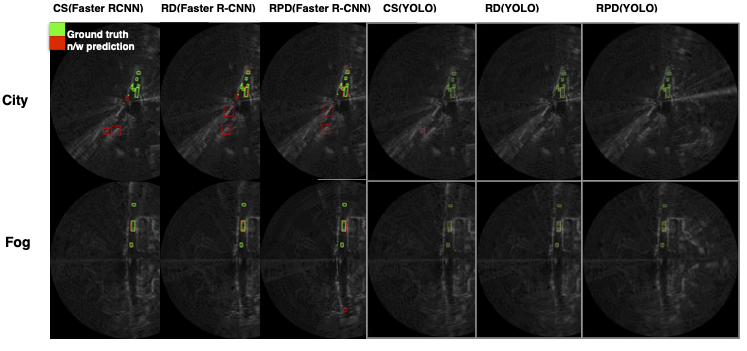}
\centering
\caption{The prediction result on radar reconstructed using the standard CS algorithm and RD, RDP tracking with Faster R-CNN and YOLO networks. The green boxes are the ground truth and the red boxes are the network predictions.}
\label{detection-results}
\end{center}
\vspace{-6mm}
\end{figure*}

\subsection{CompRPD: RADAR sub-sampling using previous RADAR}
In this paper, we experimented with the radar sub-sampling algorithm using the previous radar data on the radiate dataset. Since the radiate data was acquired in all weather conditions, similar to the previous work \cite{EUSIPCO2022}, we tested on city, motorway, snow, fog and night scenes. However, in this paper, we present the result on 40 images from each frame compared to 20 images in the previous case \cite{EUSIPCO2022}, bolstering the performance of the algorithm on additional data. We used 40\% sub-sampled anchor frames at a frequency of every 20 frames. We used Faster R-CNN and YOLOv5l models trained in accordance with the parameters reported in the previous section. In table \ref{sampling} RD refers to a baseline algorithm similar to our previous work \cite{EUSIPCO2022} but with an inverted T-shaped sampling pattern. In this paper, the proposed object detection + Kalman tracking algorithm is named RPD. Also, we compare our algorithm to the standard CS algorithm where the radar blocks were reconstructed with a uniform fixed sampling rate without prior knowledge from previous radar frames. We also test using an additional baseline algorithm CFAR. In this case, any block marked as important by the CFAR is dynamically allocated more sampling rate compared to other blocks.
At 20\% and 30\% sampling rates, the YOLO model performed better than Faster R-CNN. Also, the RD (YOLO) network was better than RPD (Faster-RCNN) at a 20\% sampling rate. Moreover, the proposed Kalman filter addition to object detection performed better than just detection-based sub-sampling at a 20\% sampling rate on both Faster R-CNN and YOLOv5 networks. However, given a 10\% sampling rate the sub-sampling algorithm using Faster R-CNN network was better than YOLOv5 and we postulate this could be because of better false positive rejections by Faster R-CNN than YOLOv5 at very low sampling rates. The overall performance is still low compared to the original at a 10\% sampling rate and for road safety, this ultra-low performing sampling rate would not be recommended for utilization. In the case of our proposed 20\% sub-sampling, RPD (YOLOv5) performs the best with a 55.6\% AP50 and 24.1\% AP. Finally, the RPD tracking algorithm is effective in improving the performance at a trade-off sampling rate of 20\%, and at higher sampling rates such as 30\% the RD results were informative for efficient sub-sampling. Across all sampling rates and networks, the previous radar-based method surpassed the standard CS and CFAR algorithms in end-to-end object detection performance. 

\begin{figure*}[ht!]
\begin{center}
\includegraphics[width=\linewidth]{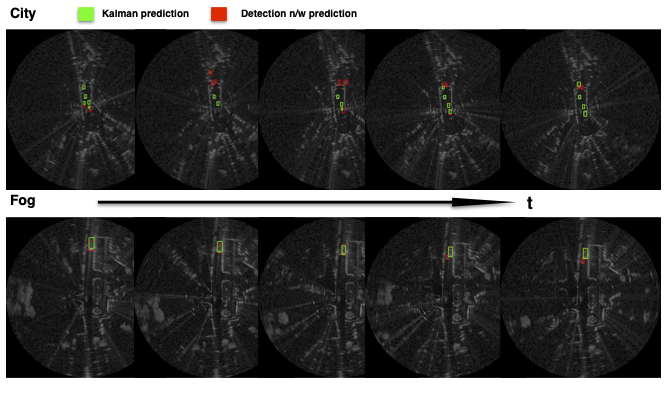}
\centering
\caption{The top row is consecutive frames from the city data set and the second row is from the fog data set. Kalman's prediction based on the previous frame is shown in green and the current frame's bounding box predictions are shown in red.}
\label{kalman-pred-detect}
\end{center}
\vspace{-6mm}
\end{figure*}

In Figure \ref{kalman-pred-detect}, we have shown the Kalman filter's prediction based on the previous frame information and the detection network's prediction on this frame across 5 frames in the city and fog scenes. 
In the case of the city scene, on frame 1, 5 objects were detected and corresponding predictions were made by the Kalman filter. However, in frame 2 the object to the right of the vehicle was not detected in frame 3, but, in the subsequent predictions, the Kalman filter's predictions were closer to the original detection's bounding box. Similarly, the object to the front (top) of the vehicle's Kalman prediction was smaller compared to the detected bounding box in frame 4 and it was immediately corrected in the next prediction in frame 5. In the case of the fog scene, the object that was detected in the 1\textsuperscript{st} frame and the subsequent Kalman filter's predictions were close to the detection n/w' prediction. Therefore, in our work, the use of the Kalman filter's predictions instead of the original network prediction from the previous frame helps in localizing the area where a higher sampling rate should be allocated. 

In Figure \ref{detection-results}, we present the detection results on radar frame reconstructed using various algorithms. In the first row, a frame from the city scene is reconstructed. Among the results reconstructed using RD (Faster R-CNN), RPD (Faster R-CNN) had the most number of predictions aligning with the ground truth predictions and it had the same number of false positives as well. However, the same image when reconstructed using the YOLO network did not have any false positives and aligned with the original ground truth boxes better compared to Faster R-CNN. 
The results on the frame from the fog scene are shown in the second row where again RD (Faster R-CNN) had more false positives than RD (YOLO).
Therefore, overall, the RDP (YOLO) algorithm has better-bounding box alignment with the ground truth and in certain cases, fewer false positive predictions as well. 

Finally, as an ablation study, to identify the importance of sampling regions blocked by the current frame in the subsequent frame for occluded object detection, we did not include these regions in our algorithm and this resulted in 46.3 AP50 and 20.3 AP on the YOLOv5l case at 20\% sampling rate. However, sub-sampling those blocked regions in our RD algorithm resulted in 54.6 AP50 and 23.8 AP. Therefore, our sub-sampling algorithm to include the surrounding regions takes occluded objects that may have been blocked by the current detected object aids in improving the detection accuracy.


\begin{table}[H]
\caption{\label{sampling}
Radar reconstruction results [AP/AP50]. Original radar AP50: 62.4 and AP: 25.3 using Yolov5l}
\centering
\begin{tabular}{|c| c| c| c|} 
 \hline
 Sampling rate & 10\% & 20\% & 30\% \\ [0.5ex] 
 \hline\hline
 Standard-CS (Faster R-CNN) & 10.0/3.6 & 37.7/15.0 & 49.2/20.3\\
 \hline
 CFAR (Faster R-CNN) &11.9/4.5 & 37.7/14.3 & 48.3/19.6\\
 \hline
 \textbf{RD} (Faster R-CNN) & 35.5/14.9  & 51.9/21.4 & 55.0/22.9\\
 \hline
 \textbf{RPD} (Faster R-CNN) & 34.3/14.7 & \textbf{53.4/21.9} & 55.0/22.8\\
 \hline
 \hline
 Standard-CS (Yolov5) & 7.1/2.4 & 36.3/14.2 & 54.3/22.0\\
 \hline
 CFAR (Yolov5) & 8.0/3.0 & 37.1/13.9 & 51.5/21.0\\
 \hline
 \textbf{RD} (Yolov5) & 27.9/12.6 & 54.6/23.8 & 61.4/26.3\\
 \hline
 \textbf{RPD} (Yolov5) & 27.9/12.7 & \textbf{55.6/24.1} & 61.4/26.3\\
 \hline
\end{tabular}
\end{table}

\subsection{Multi-rate sampling using anchor frames}
In order to deploy a sub-sampling pipeline, it is important to determine the performance vs. sampling rate trade-off in the interest of saving power while safely operating. Therefore, in Table \ref{anchor-frame}, we present the results on multiple anchor frame cases. All the frames except for the anchor frames were acquired at a 20\% sampling rate. Since we experimented on 40 images from each scene and the data is acquired at 4Hz, the in case of Multi-5, every 5\textsuperscript{th} frame was set as an anchor frame and acquired at a 40\% sampling rate. Similarly in Multi-10, every 10\textsuperscript{th} and in Multi-20, every 20\textsuperscript{th} frame was an anchor frame. Naturally, the Multi-5 has the highest AP50 performance compared to the Multi-10 and then the Multi-20. Also, for a significant increase in performance, the multi-5 is recommended compared the to Multi-20 frequency. Therefore, in safety-critical, extreme weather conditions and high population roads, overall multi-5 gives the best performance. 
As another analysis, we experimented with anchor frames acquired by direct quantization of pixels in each sample to 3 bits and dequantization. This aided in removing the need for CS to acquire the anchor frames and direct adaptive ADC is applicable in order to reduce the resolution of the ADC output across all samples. This resulted in 54.0 AP50 and 23.5 AP in the case of the Multi-20 YOLOv5 network at 20\% sampling rate, similar to the performance using CS for anchor frame acquisition. 

\begin{table}[H]
\caption{\label{anchor-frame}
Radar multi-rate sampling results [AP/AP50].}
\centering
\begin{tabular}{|c| c| c| c|} 
 \hline
 Anchor Frequency & Multi-5 & Multi-10 & Multi-20 \\ [0.5ex] 
 \hline\hline
 Faster R-CNN &  52.5/21.6  & 52.4/21.7 & 51.9/21.4\\
 \hline
 Yolov5 & 55.9/24.2  & 54.9/23.9 & 54.6/23.8\\
 \hline
\end{tabular}
\end{table}

\section{Conclusion}
In this paper, we proposed a radar sub-sampling pipeline using prior images and/or radar data and show effective reconstruction using only 10\% of the samples in case of radar and image-based algorithm and effective reconstruction using 20\% sub-sampled radar data in case of using only the previous radar-based algorithm. The previous radar-based algorithm also inherits the advantage of radar being an all-weather sensor and only relying on radar data for reconstruction. The image and radar-based algorithm was tested on the open-sourced Oxford raw radar data taken during good weather conditions and the radar-based algorithm was tested on the radiate radar data and we especially tested in fog, snow, motorway, night and city conditions. We also present the results using a hardware-efficient BPD matrix as a measurement matrix in the CS algorithm and show that it gives the best performance for the automotive case compared to BPBD and Gaussian measurement matrices. We also present the cases with multiple anchor frame frequencies and highlight the need for a higher anchor frame frequency for the best quality data acquisition and finally,
we present a YOLO-based radar-object detection network and show significant performance improvement. 
\bibliographystyle{main}
\bibliography{main.bib}
\end{document}